\pdfoutput=1
\RequirePackage{fix-cm}
\documentclass[5p,sort&compress]{elsarticle}  
\usepackage{graphicx}
\journal{}
\usepackage[numbers]{natbib}
\usepackage{graphicx}
\usepackage{amsmath,amssymb,amsfonts}
\usepackage{amsthm}
\usepackage{amsthm}
\usepackage{algorithmic}
\usepackage{color}
\usepackage[toc,page]{appendix}
\usepackage{environ}
\usepackage{array}
\usepackage{pdflscape}
\usepackage{textcomp}
\usepackage{tikz}
\usetikzlibrary{shapes.arrows}
\newcommand*\circled[1]{\tikz[baseline=(char.base)]{
		\node[shape=circle,draw,inner sep=1pt] (char) {#1};}}

\usepackage{natbib}
\usepackage{color}
\usepackage{rotate}
\usepackage{color}
\usepackage{soul}
\usepackage{listings}
\usepackage{hyperref}

\newcommand{\bea}{\begin{eqnarray*}}
	\newcommand{\eea}{\end{eqnarray*}}
\newcommand{\bne}{\begin{equation*}}

\newcommand{\ede}{\end{equation*}}

\newcommand{\bnen}{\begin{equation}}
\newcommand{\eden}{\end{equation}}
\newcommand{\bean}{\begin{eqnarray}}
\newcommand{\eean}{\end{eqnarray}}
\newcommand{\bnsn}{\begin{subequations}}
	\newcommand{\edsn}{\end{subequations}}

\newcommand{\bna}{\begin{array}}
	\newcommand{\eda}{\end{array}}
\newcommand{\bnm}{\begin{enumerate}}
	\newcommand{\edm}{\end{enumerate}}
\newcommand{\bni}{\begin{itemize}}
	\newcommand{\edi}{\end{itemize}}

\newtheorem{definition}{Definition}
\newtheorem{remark}{Remark}
\newtheorem{proposition}{Proposition}

\begin{document}

\begin{frontmatter}
	
	\title{Uninorm-like parametric activation functions for human-understandable neural models
	}
	\cortext[mycorrespondingauthor]{Corresponding author}
	
		\author[1,2,7]{Orsolya Csisz\'ar}
	\ead{csiszar.orsolya@nik.uni-obuda.hu}
	\author[2,6,7]{Luca S\'ara Pusztah\'azi \corref{mycorrespondingauthor}}
		\ead{pusztahazi.luca@uni-obuda.hu}
			\author[2,6,7]{Lehel D\'enes-Fazakas}
		\ead{denes-fazakas.lehel@uni-obuda.hu}
				\author[3]{Michael S. Gashler}
		\ead{mikegashler@gmail.com}
		\author[4]{Vladik Kreinovich}
	\ead{vladik@utep.edu}
	\author[2,5]{G\'abor Csisz\'ar}
\ead{Gabor.Csiszar@mp.imw.uni-stuttgart.de}

	\address[1]{Faculty of Electrical Engineering and Computer Science, Aalen University, Aalen, Germany}
	\address[2]{Physiological Controls Research Center, \'Obuda University, Budapest, Hungary}
	\address[4]{Department of Computer Science, University of Texas at El Paso, El Paso, TX 79968, USA}
	\address[3]{SupplyPike, Fayetteville, Arkansas 72703, USA}
	\address[5]{Institute of Materials Physics, University of Stuttgart, Stuttgart, Germany}
	\address[6]{Applied Informatics and Applied Mathematics Doctoral School, \'Obuda University, Budapest, Hungary}
	\address[7]{John von Neumann Faculty of Informatics, \'Obuda University, Budapest, Hungary}

		\begin{abstract}
	We present a deep learning model for finding human-understandable connections between input features. Our approach uses a parameterized, differentiable activation function, based on the theoretical background of nilpotent fuzzy logic and multi-criteria decision-making (MCDM). The learnable parameter has a semantic meaning indicating the level of compensation between input features. The neural network determines the parameters using gradient descent to find human-understandable relationships between input features. We demonstrate the utility and effectiveness of the model by successfully applying it to classification problems from the UCI Machine Learning Repository. 
		\end{abstract}
		
\end{frontmatter}

\section{Introduction}

There has been a great interest in combining the predictive power of neural networks with the human-under\-standable interpretations of fuzzy logic. Adaptive fuzzy systems have been studied for decades \cite{jang1991fuzzy, jang1993, Lin1996NeuralFS}  
and neural fuzzy modeling continues to be an active topic of research \cite{CHEN2013195, chen2014, kayacan2015}. 
The main benefit of combining these techniques is to provide a model with the flexibility and accuracy of black-box neural networks and the interpretability and transparency of fuzzy systems. 

In \cite{DECAMPOSSOUZA2020106275}, the authors provide a thorough review of the fundamental theories of neuro-fuzzy networks, presenting the evolution of such models over the last few decades. Recently, the variety of applications for fuzzy neural network models has increased radically. This can be seen in \cite{app12031353}, which is one of the most recent systematic reviews on the latest developments of explainable artificial intelligence (XAI), gathering the currently available approaches for adding explanations to AI/ML models and the still growing range of their application domains. One of the most powerful methods for adding explainability to neural networks is making them more like human thinking by combining them with the tools of fuzzy logic and MCDM, for which a consistent framework is given in \cite{expl}.

The main idea behind fuzzy logic is that in contrast to the traditional two-valued logic, where each statement is either true or false (represented by $1$ and $0$), we also have degrees of confidence that take intermediate values (values from the interval $[0, 1]$). In fuzzy logic,  the law of contradiction (that $x\,\&\,\neg x$ is always false) and the law of excluded middle (that $x\vee \neg x$ is always true) do not always apply. However, these laws have an intuitive appeal: The triples of ``and", ``or" and ``not" operations that satisfy these two laws are called nilpotent logical systems \cite{expl, bounded}. Here, the ``and" and ``or" operators are isomorphic to $f_{\&}(a, b)=\max (a+b-1,0)$ and $f_{\vee}(a, b)=\min (a+b,1).$  

As shown in \cite{boundedeq, boundedimpl, aggr, genbased}, the
need to be reasonably consistent with the corresponding triple affects implication operations, hedges (like ``very"), and different MCDM operations (such as preference operators).
In human thinking and decision-making, besides logical connectives, mixed operators, where a high input can compensate for a lower one, also play a significant role. 

In deep neural networks, data processing most often consists of interchangingly performing linear transformations and the element-wise rectified linear transformations $f(x)=\max (x,0)$. Interestingly, both operations $f_{\&}(a, b)=\max (a+b-1,0)$ and $f_{\vee}(a, b)=\min (a+b,1)$ and also the corresponding mixed operators can be easily represented as a composition of such neural-network transformations. As a consequence, nilpotent systems fit neural computation well. This fact helps to interpret transformations in a deep neural network in terms of these operations - and thus leads to the possibility to supplement the empirical success of deep neural networks with something which is currently largely missing in machine learning: natural-language interpretation of their results. 

Beyond neurons modelling fuzzy ``and" and ``or" operations, some of the most recent trends even involve the development of novel logic connectives, such as uninorms and nullnorms \cite{pedr_uni, pedr, hell2, hell, orient, new}. Uninorms act as conjunctions when receiving low inputs and as disjunctions when dealing with high ones. Nullnorms have the opposite behavior, that is, acting as disjunctions for low values and as conjunctions for high ones. Both uninorms and nullnorms are averaging functions when dealing with mixed inputs and as such, they play an important role in MCDM. 

Recently, in \cite{interpret}, a hybrid fuzzy neural model based on nilpotent systems was introduced, where logic gates and MCDM operators are represented by neurons with frozen weights and biases in the hidden layers. This means that the network architecture needs \textit{a priori} design. To eliminate this problem, in this study, we propose an advanced version of this model, in which the type of the logic gates and the architecture itself can be learned. The main idea is the following: instead of using frozen weights and biases in a predesigned architecture, we put the characteristic information of the logical operators in a parameterized activation function. A similar approach was introduced in \cite{param}, where the authors use an innovative, parameterized, differentiable activation function that can learn several logical operations by gradient descent. This activation function allows a neural network to determine the relationships between its input variables and provides insight into the logical significance of learned network parameters. Although they solved certain challenges with training parameterized fuzzy logic activation functions, and their model delivers human-readable output, the set of learnable logical operators is defined without a theoretical explanation. In this work, the activation functions represent uninorm-like operators with learnable parameters. These parameters have a semantic meaning: the level of compensation. Based on \cite{vlad_uni}, we know that for a specific choice of operators, weighted uninorm-based fuzzy neural networks are capable of uniformly approximating any real continuous function on a compact set to arbitrary accuracy. Also, the results in \cite{param} sum together many logical expressions, rendering them very difficult to interpret coherently. By contrast, our results are relatively simpler to interpret.

This paper is organized as follows. In Section \ref{2}, hybrid fuzzy neural models are revisited, with an emphasis on modeling conjunctions, disjunctions and uninorms using a nilpotent fuzzy framework. Section \ref{3}  revisits the concept of uninorms and the self-dual aggregative operator in nilpotent systems that play a central role in MCDM. Section \ref{4} introduces a novel approach to learning logic and MCDM operators.  Section \ref{5} demonstrates the effectiveness and utility of the model by successfully applying it to classification problems based on datasets from the UCI Machine Learning Repository. 
 Finally, Section \ref{6} summarizes the results and presents possible future  work.

\section{Neural Networks based on Nilpotent Fuzzy Logic and MCDM Tools}\label{2}
Fuzzy neural networks based on logic neurons have been used to solve a large variety of problems, such as pattern classification \cite{pattern}, time series forecasting \cite{time}, auction fraud \cite{Souza2021} and nonlinear processes \cite{nonlin}. These types of logic-based neurons are nonlinear mappings of the form $[0,1]^m\rightarrow [0,1]$ whose standard implementation of fuzzy connectives involves t-norms. More specifically, ``and" neurons are implemented through t-norms while ``or" neurons are implemented using t-conorms (s-norms). 
In \cite{whichand}, the authors analyzed which versions
of fuzzy techniques provide the best approximation to neural data processing – and, vice versa, which activation function makes the results of neural processing best describable in fuzzy terms. Interestingly, they concluded that the best activation function in this sense is exactly the ReLU function. In other words, the most popular current deep learning algorithms are also the most appropriate for fuzzy interpretation. 
In addition to ``and"-
and ``or"-operations, we also need to select a class of ``membership functions" - functions that
describe how the expert's degree of confidence that a value $x$ has some imprecise property depends on $x$. The question of which selection is the most appropriate for XAI
was also examined in \cite{XAI}.

As mentioned in the Introduction, nilpotent logical systems can model neural computation well \cite{TrillasLuk, interpret, expl}. Here, the ``and" (conjunctive) and ``or" (disjunctive) operators are isomorphic to $f_{\&}(a, b)=\max (a+b-1,0)$ and $f_{\vee}(a, b)=\min (a+b,1).$  In \cite{expl}, the authors presented a consistent theoretical framework for a hybrid neural model based on nilpotent fuzzy logic and MCDM tools. An abundant asset of operators was examined thoroughly: negations, conjunctions and disjunctions, implications, equivalence operators, and also mixed operators (where a high input can compensate for a lower one). A parametric operator was given as a generalization of nilpotent conjunctive, disjunctive, mixed  and negation operators. It was also demonstrated how this parametric operator can be applied for preference modeling. Furthermore, it was shown that membership functions, which play a substantial role in the overall performance of fuzzy representation, can also be described in this framework. Based on these results, in \cite{interpret}, the authors showed how nilpotent logical systems are suitable for neural computation and they introduced a model, where  beyond logic operators, MCDM tools can also be integrated \cite{howto}. Here, aggregation, preference and the logical operators are described by the same unary generator function, and the network architecture is designed prior to training. In the first layer of the neural network, the parameters of the membership functions are needed to be learned, while in the hidden layers the nilpotent logical operators work with given weights and biases.

This theoretical basis offers a straightforward choice of activation functions: the cutting function or its differentiable approximation, the squashing function (more details provided later) \cite{Gera, expl}. Both functions represent truth values of soft inequalities and the parameters have a semantic meaning. This model also seems to explain the great success of the rectified linear unit (ReLU). 
In \cite{why}, the authors elaborate on the empirical success of squashing functions in neural networks by showing that the formulae describing this family follow from natural symmetry requirements.

\section{Uninorms in fuzzy neural networks and the self-dual aggregative operator of nilpotent systems}\label{3}

In human thinking, besides logic operators, mixed operators, where a high input can compensate for a lower one, also play a significant role.

The concept of uninorms was introduced by Yager and Rybalov \cite{YAGER1996111}, as a generalization of both t-norms and t-conorms. Since their introduction, uninorms have been studied deeply by numerous authors from theoretical and also from application points of view. They turned out to be useful in many fields like expert systems 
and fuzzy integrals. 
\begin{definition} \cite{YAGER1996111} A mapping $U:[0,1]\times[0,1]\rightarrow[0,1]$ is a \textit{uninorm} if it is commutative, associative, nondecreasing and there exists $e\in[0,1]$ such that $U(e, x)=x$ for all $x\in[0,1].$
\end{definition}

For similar purposes, Dombi introduced the aggregative operator in \cite{duminorm}, by selecting a set of minimal concepts that must be fulfilled by an evaluation-like operator. 

\begin{definition}\cite{duminorm}
	An \textit{aggregative operator} is a function $a:[0,1]\times[0,1] \rightarrow[0,1]$ with the following properties:
	\begin{enumerate}
		\item  Continuous on $[0,1]^{2} \backslash\{(0,1),(1,0)\}$;
		\item $a(x, y)<a\left(x, y^{\prime}\right)$ if $y<y^{\prime}, x \neq 0, x \neq 1; a(x, y)<a\left(x^{\prime}, y\right)$ if $x<x^{\prime}, y \neq 0, y \neq 1$;
		\item $a(0,0)=0$ and $a(1,1)=1$ (boundary conditions);
		\item There exists a strong negation $\eta$ such that $a(x, y)=\eta(a(\eta(x), \eta(y)))$ (the self-De Morgan identity) \\if $\{x, y\} \neq\{0,1\}$;
		\item $a(1,0)=a(0,1)=0$ or $a(1,0)=a(0,1)=1$.
	\end{enumerate}
	
\end{definition} 

The main difference in the definitions of the uninorms and aggregative operators is that the self-duality requirement does not appear in uninorms, and the neutral element property is not in the definition for the aggregative operators.

Some of the most recent trends in the neuro-fuzzy systems involve the development of new operators and logic connectives, including uninorms \cite{new, advanced, DECAMPOSSOUZA}. Traditional logic neurons use t-norms and s-norms to perform the operations. Recently, several works have addressed uninorm-based neurons considering uninorms to extend classic logic neurons models by replacing the t-norms with uninorms. The neurons constructed from uninorms are called unineurons \cite{pedr_uni}. 
As a remarkable feature, these unineurons implement an important characteristic of biological neurons, neuronal plasticity. Throughout learning, the internal neuron processing function can be modified in response to external changes. 
Since they are a generalization of ``and" and ``or" neurons, they are more flexible and capable of handling complex logical rules. Experimental results suggest that fuzzy neural networks created from uninorms or nullnorms give accurate results, perform better and provide more human-understandable information to extract.  

Next, we recall the concept of the uninorm-like weighted general operators  in nilpotent systems \cite{aggr}, which are included in our fuzzy neural model. 

To make the notations more readable, we use the following definition:
\begin{definition}
	Let us define the \textit{cutting function}, denoted by $[ \;]$ as
	\begin{equation}\label{eq:cut}
		[x] = \begin{cases}
			0,\quad if \qquad x<0,\\
			x, \quad if \quad 0\leq x \leq 1,\\
			1, \quad if \qquad 1<x.
		\end{cases}
	\end{equation} 
\end{definition}

\begin{remark}
Note that using the cutting function, $f_{\&}(a, b)=\max (a+b-1,0)=\left[a+b-1\right]$ and $f_{\vee}(a, b)=\min (a+b,1)=\left[a+b\right].$ 
\end{remark}

\begin{definition}
Let $\mathbf{w}=\left(w_{1}, \ldots, w_{n}\right)$ and $w_{i}>0$ be real parameters, $f:[0,1] \rightarrow[0,1]$ a monotonically increasing bijection with $\nu \in[0,1]$. The \textit{weighted general operator} is defined by
\begin{equation}\label{operator}
a_{\nu, \mathbf{w}}(\mathbf{x}):=f^{-1}\left(\left[\sum_{i=1}^{n} w_{i}\left(f\left(x_{i}\right)-f(\nu)\right)+f(\nu)\right]\right) .
\end{equation}
\end{definition}

\begin{proposition}
The weighted general operator in \eqref{operator} includes the following special cases:
\begin{enumerate}
	\item For $f(\nu)=1$ and $w_{i}=1\ \text{for all} i$, it is a conjunction with the generator function $1-f(x)$.
	\item For $f(\nu)=0$ and $w_{i}=1\ \text{for all} i$, it is a disjunction with the generator function $f(x)$.
	\item For $f(\nu)=\frac{1}{2}$ or $\sum_{i=1}^{n} w_{i}=1$, it satisfies the self-De Morgan property.
	\item For $f(\nu)=\frac{1}{2}$ and $w_{1}=\cdots=w_{n}$, or for $w_{i}=\frac{1}{n}$, it is a weighted aggregative operator (a commutative self-De Morgan operator).
	\item For $\nu=0$ or $\sum_{i=1}^{n} w_{i} \geq 1$, it satisfies the boundary condition $a_{\nu}(\mathbf{0})=0$.
	\item For $\nu=1$ or $\sum_{i=1}^{n} w_{i} \geq 1$, it satisfies the boundary condition $a_{\nu}(\mathbf{1})=1$.
	\item In particular for two variables, with $\frac{1}{2} \leq w_{1}=w_{2}$ and $f(\nu)=\frac{1}{2}$, it is

-- commutative,

-- satisfies the De Morgan property,

-- satisfies the boundary conditions (i.e. $a(0,0)=0$ and $a(1,1)=1$ ),

-- $a(0,1)=a(1,0)=\nu$,

-- if $x, y \leq \nu$, then $a(x, y) \leq \nu$,

-- if $x, y \geq \nu$, then $a(x, y) \geq \nu$.
	\item For two variables, with $w_{1}=w_{2}=1$ and $f(\nu)=\frac{1}{2}$, it is

-- commutative,

-- satisfies the De Morgan property,

-- satisfies the boundary conditions $($ i.e. $a(0,0)=0$ and $a(1,1)=1)$,

-- $a(0,1)=a(1,0)=\nu$,

-- it is conjunctive for $x, y \leq \nu$,

-- it is disjunctive for $x, y \geq \nu$,

-- it is averaging for $x \leq \nu \leq y$ and for $y \leq \nu \leq x$.
	\item For one variable and with $w=-1$, it is a negation operator with generator $f(x)$.
	\item $a_{w}(n(x), y)=f^{-1}\left[w(f(y)-f(x))+\frac{1}{2}\right]=p_{w}(x, y)$. 
\end{enumerate}
\end{proposition}
Note that for $f(x)=x$, \eqref{operator} gives the functions modelled by neurons in neural networks:
\begin{equation}\label{nn}
\left[\sum_{i=1}^{n} w_i x_i+C\right].
\end{equation}	

\begin{remark}
Using the weighted general operator \eqref{operator}, we can define disjunctive, conjunctive and aggregative operators which differ only in one parameter (see Table \ref{tab:operators}). Here, the parameter has an important semantic meaning as the level of compensation, being minimal in the case of the conjunction, and maximal in the case of the disjunction.
	
\end{remark}

In the following text, we will implement \eqref{operator} as a two-variable operator where the weights are constant and $f$ is the identity, i.e.
\begin{itemize}
\vspace{-3pt}	\item $ \mathbf{x} = (x,y)$,
\vspace{-3pt}	\item $ \mathbf{w} = (w_1, w_2) = (1,1)$,
\vspace{-3pt}	\item $ f(x) = x$,
\end{itemize}
\vspace{-3pt}
yielding \begin{equation}\label{anu}
 a_\nu(\mathbf{x}) = [x+y-\nu].
\end{equation}

To be consistent with \cite{param}, in the following we will use the notation $\alpha := \nu$ and 
$x\textcircled{$\alpha$}y := a_\nu(\mathbf{x}) $ that gives us 
\begin{equation}
	x\circled{$\alpha$} y = [x+y-\alpha].
\end{equation}
In the formula of the weighted general operator, there appears the cutting function which is not differentiable by definition. However, the fuzzy neural network often needs to be able to learn its parameters by a gradient based optimization method, therefore it is preferable for the operator to be everywhere differentiable. For this reason an approximation of the cutting function should be used. The so-called squashing function has many ideal properties for this purpose \cite{Gera, expl}.

\begin{definition}
	By a \textit{squashing function}, we mean the following function
	\begin{equation}\label{eq:sq}
	S_{a, \lambda}^{(\beta)} (x) = \frac{1}{\lambda\beta} \ln \frac{1+e^{\beta (x-(a-\lambda/2))}}{1+e^{\beta (x-(a+\lambda/2))}}
	\end{equation} 
\end{definition}

\begin{figure}[h!]
\centering
	\caption{Squashing function with $\beta=10$}
	\vspace{0.15cm}
\includegraphics[width=0.48\textwidth]{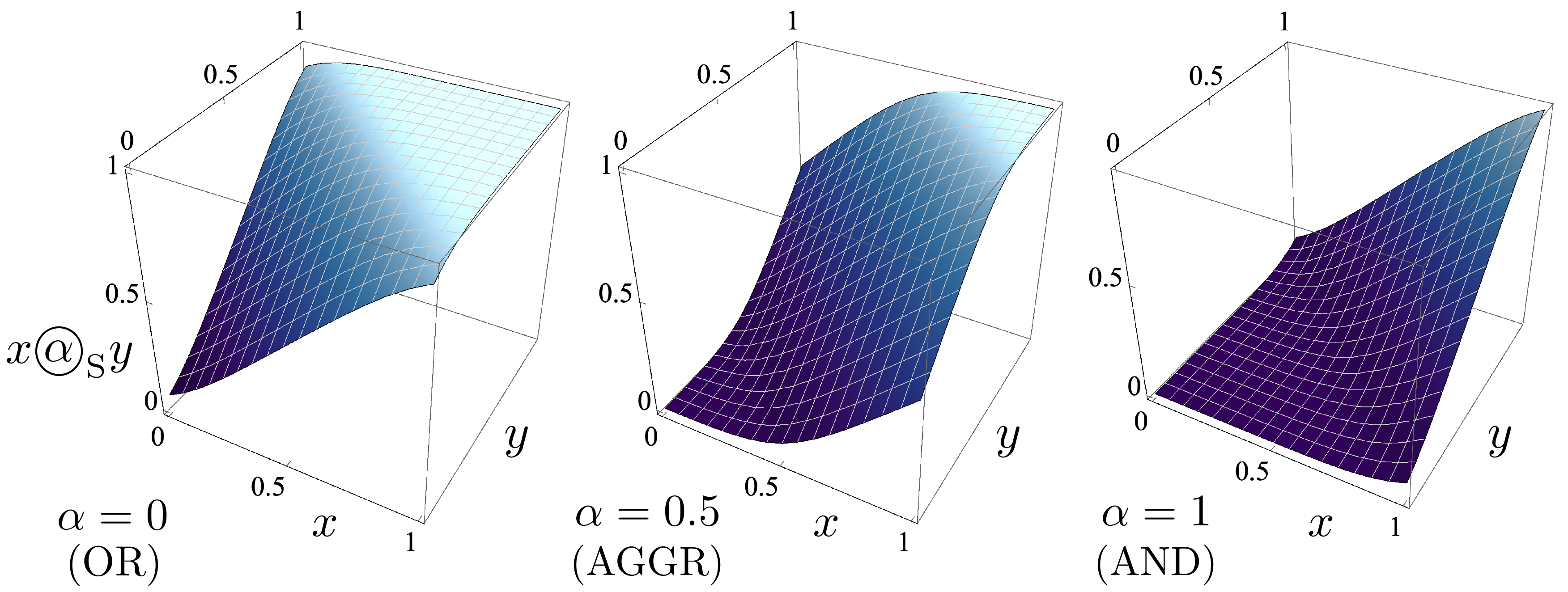}
\label{fig:sq10}
\end{figure}

\begin{figure}[h!] 
\centering
	\caption{Squashing function with $\beta=50$}
	\vspace{0.15cm}
\includegraphics[width=0.48\textwidth]{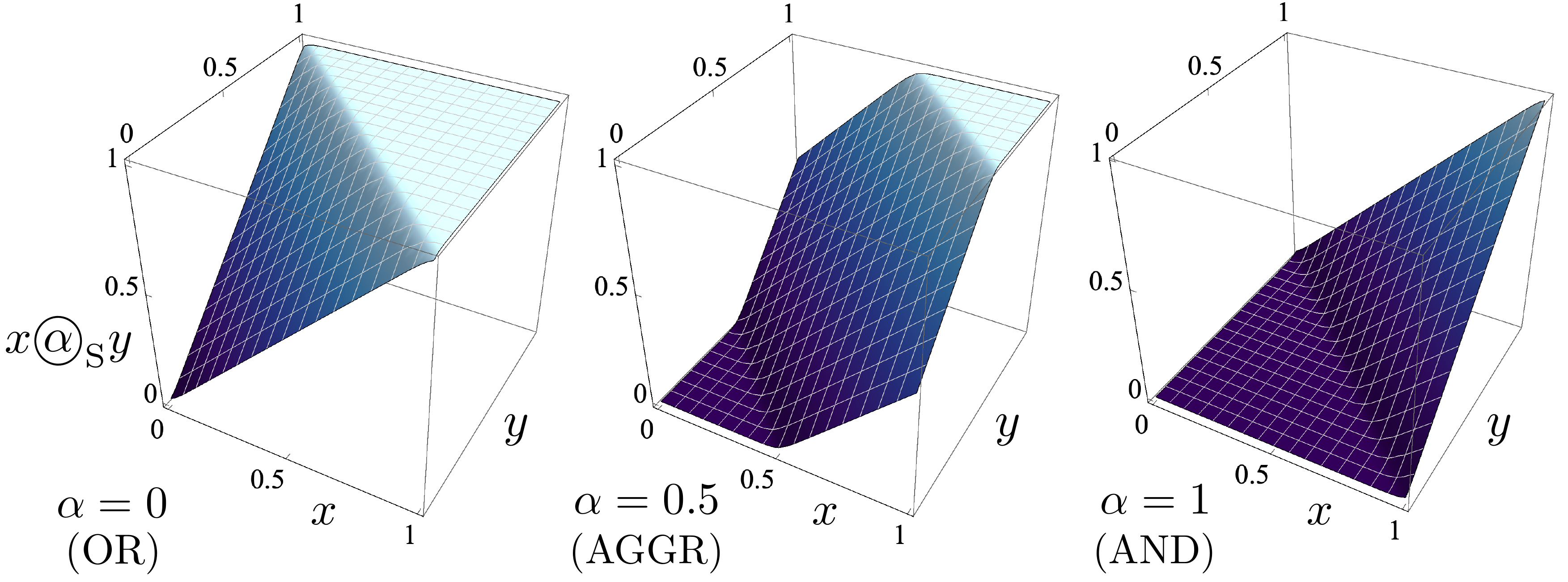}
\label{fig:sq50}
\end{figure}

It can be proven that as $\beta \rightarrow \infty$ the squashing function converges to the cutting function, hence we can use it as a differentiable approximation of the non-smooth cutting function. In Figures \ref{fig:sq10} and \ref{fig:sq50}, two visual representations of the operators using the squashing function can be seen. 

\vspace{-0.9cm}
In the implementation in Section \ref{5}, we will use the following parameter values for the squashing function: $\lambda = 1$, $a = \frac{1}{2}$, $\beta = 80$, which gives us
\begin{equation}\label{S} S(x) = \frac{1}{80} \ln \frac{1+e^{80 x}}{1+e^{80 (x-1)}}.
\end{equation}

For now, the parameters are chosen intuitively for our purposes in this validation, however, their further tuning can be a possible next step.

Approximating the cutting function in \eqref{nn} by $S$ in \eqref{S}, we get
\begin{equation}
	\left[\sum_{i=1}^{n} w_i x_i+C\right]\approx S\left(w_i x_i+C\right).
\end{equation}	

In particular, for $n=2, w_i=1$, we get a parametrized operator, which  can be disjunctive, conjunctive or aggregative depending on the value of parameter $C$ (see Table \ref{tab:operators}).

\begin{table}[h!]
	\begin{center}
		\caption{Special cases of the weighted general operator}
		\label{tab:operators}
		\begin{tabular}{|l|c|r|r|} 
			\hline 
			& $\alpha$ & $x\circled{$\alpha$} y$ & $x\circled{$\alpha$}_S y$\\
			\hline
			Disjunction & $0$ & $[x+y]$ & $S(x+y)$  \\
			\hline
			Conjunction & $1$ &$[x+y-1]$ & $S(x+y-1)$ \\
			\hline
			Aggregative op. & $0.5$ &$[x+y-0.5]$ & $S(x+y-0.5)$ \\
			\hline
		\end{tabular}
	\end{center}
\end{table}

Thus, the weighted general operator we use provides a general framework for different types of fuzzy logic operators. As shown in \cite{aggr}, based on the parameters, conjunctive, disjunctive, self-dual and even a negation operator can be obtained.

\section{Approach to Learning Logic and MCDM Operations}\label{4}

In \cite{param}, the authors introduced a novel approach using an adaptive activation function that is able to learn several logical operations. Already, there exist models which are going towards better interpretability through the weights used, however, this model has the advantage of being able to learn which logical operator is the optimal choice to use. This unique activation function is parameterized, continuous and differentiable, therefore it can be trained to choose between different logical expressions by gradient descent, instead of setting by hand.

The activation function of the neural network in \cite{param} is 
$$x \textcircled{$\alpha$} y = \frac{(x+\alpha)(y+\alpha)}{|\alpha|+1}-|\alpha|,$$ where $x, y, \alpha \in [-1,1]$.
By tuning the parameter $\alpha$ in this formula, several logical expression can be obtained, e.g.
\begin{itemize}
\vspace{-3pt}	\item $\mathbf{identity}(x) = \mathbf{true} \circled{0} x$,
\vspace{-3pt}	\item $\mathbf{not}(x) = \mathbf{false} \circled{0} x$,
\vspace{-3pt}	\item $\mathbf{and}(x,y) = x \circled{1} y$.
\end{itemize}
\vspace{-3pt}

Our model is based on the previously introduced idea of using a parameterized, trainable activation function. The main difference is in the applied function. Although the model in \cite{param} delivers a human-readable output, the set of learnable logical operators is defined without a theoretical background. Further, their model summed together many unrelated logicl expressions, rendering the aggregate expressions counter-intuitive to interpret. In this work, we apply a unified framework using the weighted general operator \eqref{operator}. For different parameter values, it can play the role of a disjunction, a conjunction, and a uninorm-like aggregative operator (see Table \ref{tab:operators}). 
The aggregative operator acts as a conjunction when receiving low inputs and as a disjunction when dealing with high ones. When dealing with mixed inputs, it has an averaging property. 

In our implementation, the adaptive activation function is defined by the formula
$$x\circled{$\alpha$}_S y = S(x+y-\alpha),$$ 
where $x,y,\alpha \in [0,1]$ (for details see Section \ref{3}). As mentioned before, the cutting function \eqref{eq:cut} is not differentiable, therefore we use its differentiable approximation, the squashing function \eqref{eq:sq}. In this neural network, we use the innovative idea of \cite{param}, namely that the optimizer is not applied on the weights of the aggregation functions but on the adaptive activation function, through which the appropriate logical or MCDM operator is learned.

\section{Results and Discussion}\label{5}

The main purpose of this research project is to introduce the mathematical framework for using the weighted general operator as the activation function in fuzzy neural networks. The next step is to apply it to real-world problems, which are restricted to classification problems in this article. The model from \cite{param} gives an appropriate candidate for our goals, therefore with some necessary modifications, the fuzzy neural network is as follows: 

The neural network consists of $9$ layers: first, the inputs are fed into a normalization layer, which maps the inputs onto the interval $[-1,1]$. Then they feed into an AllPairings layer, which outputs each pairing of the input values, and also pairs each value with True and False, i.e. with $1$ and $-1$. The next layer is called the FuzzyLogic layer, which learns the optimal fuzzy logic operation for each pairing through the activation function. Compared to \cite{param}, this activation function has been changed to the (simplified) weighted general operator. Next, a FeatureSelector layer is applied to manage the dimensionality. The last three layers together can be referred to as the logic part of the neural network. The depth of the neural network can be set by increasing or decreasing the number of logic parts, for our experiments, it is fixed to two. Between two logic parts, there needs to be a layer to remap the previous outputs onto the interval $[-1,1]$ using the $tanh$ activation function. After the second logic part, the $9^{th}$ layer is a $max$ layer for classification. 

As one can see, the neural network has inputs and outputs from the interval $[-1,1]$, while the domain and the range of the squashing function is the interval $[0,1]$. In order to  get the proper inputs for all functions in the implementation, two transformations were applied of the forms 


\begin{itemize}
\vspace{-3pt}	\item $z \mapsto \tilde z = 2z-1, \; \qquad z \in [0,1], \tilde z \in [-1,1]$,
\vspace{-3pt}	\item $\tilde z \mapsto z = (\tilde z+1)/2, \quad z \in [0,1], \tilde z \in [-1,1]$.
\end{itemize}
\vspace{-3pt}

Our model was applied to 12 classification problems from the UCI Machine Learning Repository \cite{Dua:2019}.
The results show a great variety in the mean error as well as in the nearest approximating logic. Since the goal of this work is to implement an explainable neural network, if the resulting logical expression is too long to be understandable or contains just a logic constant, then it cannot be considered a proper output, hence they were omitted from tables \ref{tab:error_rates} and \ref{tab:logic}. However, we do not claim that our model only yields short (but not too short) and explainable results. It is planned as future work to improve the model by refining the fuzzy layer together with the descriptive function in order to get more human-readable results.

For the validation, the learning rate is fixed to $0.01$, and the regularization term is fixed to $0.0001$. As a reference model, we use a deep neural network with a topology (depth, number of weights) similar to the fuzzy model, although all the layers are fully connected. The activation function is the $tanh$ function throughout the DNN. In Table \ref{tab:error_rates}, the resulting misclassification rates of the previously described DNN and the fuzzy neural network are listed. The misclassification rates are similar in most of the cases, however, the fuzzy model has the benefit of better interpretability. The corresponding nearest approximating logic can be found in Table \ref{tab:logic}. The operation $uni$ in the expressions is the uninorm-like aggregative operator 
$$x\circled{$0.5$}_S y = S(x+y-0.5).$$ The output is to be interpreted as a continuous value between $0$ and $1$. For the $2$-class classifications, the threshold value is $0.5$.

\begin{table}[h!]
	\begin{center}
		\caption{Misclassification rates}
		\label {tab:error_rates} 
		\begin{tabular}{|l|c|c|} 
			\hline 
			Dataset &  DNN & Fuzzy NN\\
			\hline
			Breast cancer & $0.23$ & $0.25 $ \\
			\hline
			Diabetes & $0.26$ & $0.28$ \\
			\hline
			King-Rook vs King-Pawn & $0.06$ & $0.07$ \\
			\hline 
			Vote & $0.05$ & $0.29$ \\
			\hline
		\end{tabular}
	\end{center}
\end{table}

\begin{table}[h!]
	\begin{center}
		\caption{Nearest approximating expression}
		\label{tab:logic}
		\begin{tabular}{|l|c|} 
			\hline 
			Dataset & Expression \\
			\hline
			Breast cancer & $((28) uni (34)) uni ((6) uni (34))$\\
			\hline
			Diabetes & $ 1-(
			(1) uni (6) )$\\
			\hline
			King-Rook vs King-Pawn &  $((9) uni (34)) uni ((22) uni (34)) $ \\
			\hline
			Vote & $((25) uni (31)) uni ((11) uni (37)) $ \\
			\hline 
		\end{tabular}
	\end{center}
\end{table}

\section{Conclusions}\label{6}
We presented a deep learning model for finding human-understandable connections between input features. Our approach uses a parametrized, differentiable activation function, based on the so-called weighted general operator of nilpotent logical systems. By changing the parameter value, conjunctive, disjunctive and mixed operators can be obtained, where the parameter has a semantic meaning indicating the level of compensation between input features. We approximated the cutting function in the formulae for the sake of differentiability. The neural network determines the parameters using gradient descent and finds human-understandable relationships between the input features. We demonstrated the utility and effectiveness of the model by successfully applying it to classification problems from the UCI Machine Learning Repository.  

Our model uses only three logical and MCDM operators, which can be extended in the future. With a more diverse list of learnable operations, more complex fuzzy decisions can be formulated. Adjusting the parameter $\beta$,``the smoothness factor", of the squashing function, also remains for future work. Moreover, fuzzifying the descriptive function to simplify the resulting logical expression should also be considered a future improvement of the model. Although the results are not yet the easiest to interpret, this work demonstrates a promising step towards better explainability of fuzzy deep learning models. We are working on the implementation and testing of the python code, which will be available on the $\dots$ website.

Intuitively, we want an operation from $[0,1] \times [0,1]$ to $[0,1]$ which is the fastest to compute. In a computer, in effect, the only hardware supported operations are addition and multiplication, with multiplication being much slower than addition. Thus, it is better to restrict ourselves to addition. With a single addition, the only operation we can perform on two inputs that does not ignore one of them is $a  + b$. With two additions, we can have $2a+b$ or $a+2b$ or $a + b + c$. The last one is the operation proposed in this work. Cutting, on the other hand, is automatically done by computers, we just restrict it to the interval $[0,1] $ instead of cutting between $[- \infty, +\infty]$, where the computer automatically takes all values larger than some threshold as $+\infty$ and similarly, takes all values smaller than some threshold as $-\infty$.

\section*{Acknowledgements}

The project no. 2019-2.1.11-TÉT-2020-00217 has been implemented with the support provided by the National Research, Development and Innovation Fund of Hungary, financed under the 2019-2.1.11-TÉT-2020-00217 funding scheme. Project no. 2019-1.3.1-KK-2019-00007. has been implemented with the support provided from the National Research, Development and Innovation Fund of Hungary, financed under the 2019-1.3.1-KK funding scheme. This project has been supported by the National Research, Development, and Innovation Fund of Hungary, financed under the TKP2021-NKTA-36 funding scheme. L. Dénes-Fazakas was supported by the ÚNKP-21-3 New National Excellence Program of the Ministry for Innovation and Technology.

\bibliographystyle{abbrv}
\bibliography{uninorm_activation}


\end{document}